\documentclass[10pt,twocolumn,letterpaper]{article}

\usepackage{cvpr}
\usepackage{times}
\usepackage{epsfig}
\usepackage{graphicx}
\usepackage{amsmath}
\usepackage{amssymb}
\usepackage{float}
\usepackage{arydshln}
\usepackage{placeins}
\graphicspath{{graphs/}}

\usepackage[pagebackref=true,breaklinks=true,letterpaper=true,colorlinks,bookmarks=false]{hyperref}

\cvprfinalcopy 

\def\cvprPaperID{9951} 

\ifcvprfinal\fi
\begin{document}

\title{Homography Estimation with Convolutional Neural Networks Under Conditions of Variance}

\author{David K. Niblick\\
Purdue University\\
610 Purdue Mall, West Lafayette, IN 47907\\
{\tt\small dniblick@purdue.edu}
\and
Avinash Kak\\
Purdue University\\
610 Purdue Mall, West Lafayette, IN 47907\\
{\tt\small kak@purdue.edu}
}

\maketitle

\begin{abstract}
Planar homography estimation is foundational to many
computer vision problems, such as Simultaneous Localization
and Mapping (SLAM) and Augmented Reality (AR). However,
conditions of high variance confound even the
state-of-the-art algorithms. In this report, we analyze the
performance of two recently published methods using
Convolutional Neural Networks (CNNs) that are meant to
replace the more traditional feature-matching based
approaches to the estimation of homography.  Our evaluation
of the CNN based methods focuses particularly on measuring
the performance under conditions of significant noise,
illumination shift, and occlusion. We also measure the
benefits of training CNNs to varying degrees of noise.
Additionally, we compare the effect of using color images
instead of grayscale images for inputs to CNNs. Finally, we
compare the results against baseline feature-matching based
homography estimation methods using SIFT, SURF, and ORB. We
find that CNNs can be trained to be more robust against
noise, but at a small cost to accuracy in the noiseless
case.  Additionally, CNNs perform significantly better in
conditions of extreme variance than their feature-matching
based counterparts.  With regard to color inputs, we
conclude that with no change in the CNN architecture to take
advantage of the additional information in the color planes,
the difference in performance using color inputs or
grayscale inputs is negligible.  About the CNNs trained with
noise-corrupted inputs, we show that training a CNN to a
specific magnitude of noise leads to a ``Goldilocks Zone''
with regard to the noise levels where that CNN performs
best.
\end{abstract}

\section{Introduction}


A homography is a planar projective transformation
represented by a $3 \times 3$ non-singular matrix in the
homogeneous coordinate system. The matrix $\textbf{H}$ maps
homogeneous coordinates $\textbf{x}$ to $\textbf{x}'$. This
transformation is foundational to many computer vision
problems, including Simultaneous Localization and Mapping
(SLAM), 3D reconstruction, Augmented Reality (AR), etc.
\cite{Dubrofsky09homographyestimation}
\cite{Vincent2001DetectingPH} Failing to accurately estimate
a homography, especially in non-ideal conditions for
automated applications, can cause significant error that
propagates through the system. For example, in AR
applications that require tracking the pose of a camera,
even ``minor'' camera motions create estimation error that
cause state-of-the-art systems to fail \cite{SLAM}.

The baseline homography estimation algorithms depend on
detecting features in a pair of images, determining
candidate feature correspondences by comparing the
descriptor vectors associated with the features, eliminating
the outliers in the candidate correspondences with RANSAC
\cite{ransac}, and, finally, using a nonlinear least-squares
method (like the Levenberg-Marquardt algorithm) over the
inlier set to estimate the homography.  Commonly used
algorithms for feature detection and description include
Scale Invariant Feature Transform (SIFT) \cite{sift},
Speeded Up Robust Features (SURF) \cite{surf}, and Oriented
FAST and Rotated BRIEF (ORB) \cite{orb}. While these
algorithms have been used in various applications for over a
decade, they all produce inaccurate results under non-ideal
conditions. Noise, illumination variance, and obscuration
are common occurrences in many automated computer vision
applications that are vulnerable to the erroneous
estimations of homography as produced by the
feature-matching based methods.

The recent success of Convolutional Neural Networks (CNNs)
in computer vision applications has inspired research in
using deep learning for homography estimation.  In computer
vision, CNNs have achieved immense success at object
detection and classification, image synthesis, and stereo
matching \cite{Simonyan2014VeryDC}
\cite{goodfellow2014generative}
\cite{han2015matchnet}. Against these advances in
deep-learning based solutions to computer vision problems,
this report presents an exhaustive evaluation of two CNN
based tools for estimating homographies
\cite{DBLP:journals/corr/DeToneMR16} \cite{Nowruzi}.  The
authors of these two networks have reported significant
improvements over what can be achieved with the traditional
homography estimation methods using the ORB \cite{orb}
features.\footnote{It is surprising that the authors of
  neither \cite{DBLP:journals/corr/DeToneMR16} nor
  \cite{Nowruzi} compared the performance of their CNN based
  homography estimators with what can be achieved with SIFT
  and SURF --- two very commonly used operators used in
  calculating the homography between two images.  A
  comparison of the CNN-based homography estimation with the
  SIFT-based approach has been reported in
  \cite{DBLP:journals/corr/abs-1709-03966}.  With regard to
  a comparison of CNN-based homography estimators with the
  feature-matching based methods, our report here expands on
  the work in \cite{DBLP:journals/corr/abs-1709-03966}
  through a more exhaustive comparison that not only
  includes varying degrees of noise but also varying degrees
  of illumination and obscuration effects.}

The CNN-based homograpahy estimation solution presented in
\cite{DBLP:journals/corr/DeToneMR16} uses an altered version
of the VGG-14 \cite{Lin2014MicrosoftCC} architecture to
calculate the four point parameterization of the homography
matrix between an image and a warped image. They used the
Common Objects in Context (COCO) Dataset
\cite{Lin2014MicrosoftCC} for training and testing. A key
part of their method is the process they employed for
generating the training data, in which they warped an image
with a randomly generated homography, and then passed that
homography as the ground truth. The reported results were
approximately 20\% more accurate than those based on the
feature-matching method using ORB.

The second CNN-based homograpahy estimation solution,
presented in \cite{Nowruzi}, uses multiple Siamese CNNs in a
sequence with an attempt to achieve accuracies better than
those reported in \cite{DBLP:journals/corr/DeToneMR16}. With
the ``boosting-like'' effect achieved by their architecture,
they claimed a 67\% error reduction compared to what could
be achieved with the feature-based method using ORB.

Our main focus in this paper is the analysis of the performance
of the CNN based approaches for homography estimation under
conditions of high variance --- {\em that is, when the image
  data is corrupted with varying amounts of noise, with
  varying degrees of illumination changes, and with varying
  degrees of occlusion.}  In addition to comparing the two
CNN based approaches with each other, we also report on
their comparison with the baseline methods based on SIFT,
SURF, and ORB.

With regard to noise, of particular significance in our
comparative evaluations is the effect on performance when
the CNNs are trained on noise-corrupted data vis-a-vis the
case when they are trained on just clean data.  When the
CNNs are trained with noise-corrupted inputs, we show that
training a CNN at a specific level of noise leads to a
``Goldilocks Zone'' with regard to the noise level where
that CNN performs best.  That is, the CNN-based homography
estimator produces the best average performance when the
noise level (as characterized by its statistical properties)
that was used during the training matches the noise level in
the images on which the CNN is tested.

Our comparative evaluation also includes results on color
images. That part of the evaluation has required modifying
slightly the CNNs proposed in
\cite{DBLP:journals/corr/DeToneMR16} and \cite{Nowruzi}
since now we must account for the three color planes at the
input to the CNNs.  For the case of color, we show that,
with the same overall CNN architecture, the additional
information in the color planes makes negligible
contributions to the accuracies of the estimated
homographies, implying that, perhaps, with additional
channels in the convolutional layers --- especially those
that are closer to the input --- one might be able to
enhance the accuracies.

All our tests were conducted on the Open 2019 Dataset
\cite{OpenImages} \cite{OpenImages2}, which is the first
attempt at using this dataset for homography estimation with
CNNs. We find that, although SIFT is more accurate under
ideal conditions, CNNs perform significantly better under
conditions of variance, especially with regard to additive
noise. We also show that the performance of a CNN-based
homography estimator for noisy images can be improved if the
CNN is trained with noisy training data containing roughly the
same level of noise as is expected in the test data. This
improvement vis-a-vis the noise comes at a small cost to the
performance for the noiseless case.  The key takeaway from
our paper is that no one method should be treated as
``universally superior'' to all others, but instead
environmental conditions and engineering constraints need to
be deliberately considered when choosing the best method for
any application.

In the rest of this paper, we start in Section
\ref{sec:cnn_for_homography} with a review of the two
CNN-based homography estimators mentioned above.  We talk
about the datasets and the metrics used in our comparative
evaluation in Section \ref{sec:datasets_used}.  In Section
\ref{sec:comparative_eval}, we describe the experiments we
have carried out for meeting the goals of this paper.
Section \ref{sec:results} presents the results of our
comparative evaluation.  Finally, we conclude in Section
\ref{sec:conclusion}.

\section{Homography Estimation With CNNs}
\label{sec:cnn_for_homography}

In this section, we provide a brief review of the two
CNN-based approaches to homography estimation that we have
used in our comparative evaluation.


\subsection{Deep Image Homography Estimation (DH)}
\label{sec:dhcnn}

\begin{figure*}
\vspace{-0.1in}
\begin{center}
\includegraphics[width=0.8\linewidth,height=3.5cm,keepaspectratio]{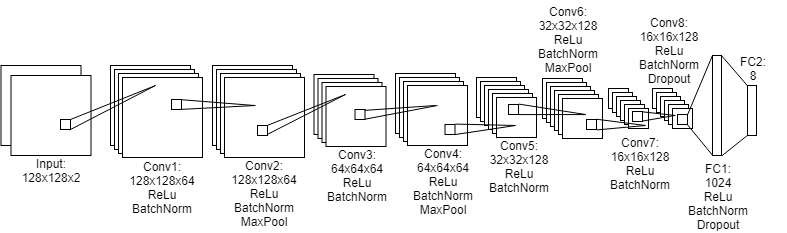}
\end{center}
   \caption{Deep Homography Estimation Model \cite{DBLP:journals/corr/DeToneMR16}}
\label{fig:dh_model}
\vspace{-0.2in}
\end{figure*}
The CNN-based approach to homography estimation as proposed
in \cite{DBLP:journals/corr/DeToneMR16} uses the VGG-14 Net
\cite{Simonyan2014VeryDC} to process two patches from an
image pair. The two input patches are converted to
grayscale, normalized, and then are stacked onto each
other. The architecture uses eight convolutional layers,
followed by two fully connected layers (Figure
\ref{fig:dh_model}). The convolutional layers use a $3
\times 3$ kernel. ReLu activation and batch normalization
follows each layer. Max Pooling occurs prior to the third,
fifth, and seventh convolutional layers. Dropout with a
factor of 0.5 is applied before each fully connected layer.

The output of the neural network is a parameterized form of
the homography matrix --- the parameterized form is known as
the four-point parameterization and denoted
$\textbf{H}_{4pt}$. To avoid confusion in the rest of the
paper, we will use $\textbf{H}_{mat}$ to denote the original
homography matrix and $\textbf{H}_{4pt}$ its four-point
parameterization.  Using homogeneous coordinates, if
$\textbf{x}$ is a point in one image and $\textbf{x}'$ the
corresponding point in the other image, the two are related
by $\textbf{x}' = \textbf{H}_{mat} \textbf{x}$. If the 
relationship between homogeneous coordinates and the actual
pixel coordinates is given by
$\begin{bmatrix} x_1 & x_2 & x_3
\end{bmatrix} 
=
\begin{bmatrix}
    u  &
    v  &
    1  
\end{bmatrix}$
, we can write the following form for the matrix
$\textbf{H}_{4pt}$:

\begin{equation}
\textbf{H}_{4pt}  = 
\begin{bmatrix}
u_1 - u_1' & v_1 - v_1' \\
u_2 - u_2' & v_2 - v_2' \\
u_3 - u_3' & v_3 - v_3' \\
u_4 - u_4' & v_4 - v_4'
\end{bmatrix}
\label{eq:eight_point}
\end{equation}

\noindent where $(u_i,v_i), i=1,2,3,4$ are the pixel
coordinates of some four corners in one image and
$(u_i',v_i'), i=1,2,3,4$ the corresponding four corners in
the other image.  By adding the elements of
$\textbf{H}_{4pt}$ to the coordinates of the corners in the
first image of an image pair, we obtain the coordinates of
the corresponding points in the second image.  We can
subsequently use the pixel coordinates of the four
corresponding corners to calculate $\textbf{H}_{mat}$ from
$\textbf{H}_{4pt}$.

The training of the CNN in
\cite{DBLP:journals/corr/DeToneMR16} is done through a
self-supervised method in which a patch is randomly selected
from the image, no closer than $\rho$ pixels from any image
edge. A random perturbation (of no more than $\rho$ pixels)
is applied to the corners of the patch, creating warped
corners. The correspondences between the original corners
and warped corners are used to calculate a homography,
$\textbf{H}_{mat}$. The homography is applied to the image,
and a patch (from same coordinates as original patch) is
extracted from the warped image as the warped patch. The
original and warped patches are stacked and input into the
neural network, with the corner perturbation values used as
the ground truth, $\textbf{H}_{4pt}^{target}$. This ``label
on the fly'' method is both simple and flexible. The ability
to use any image dataset for training and testing makes this
a powerful approach for homography estimation.

\subsection{Homography Estimation with Hierarchical Convolutional Networks (HH)}
\label{sec:hhcnn}

\begin{figure*}
\vspace{-0.1in}
\begin{center}
\includegraphics[width=0.8\linewidth,height=6cm,keepaspectratio]{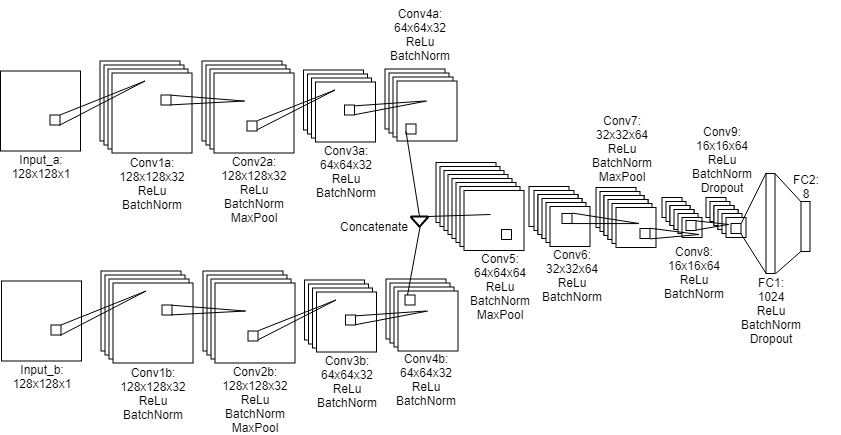}
\end{center}
   \caption{Hierarchical Homography Model \cite{Nowruzi}}
\label{fig:hh_model}
\vspace{-0.1in}
\end{figure*}

\begin{figure*}
\begin{center}
\includegraphics[width=0.8\linewidth, height=3cm,keepaspectratio]{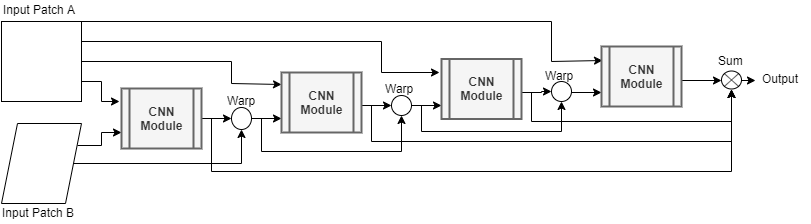}
\end{center}
   \caption{Hierarchical Homography Stack \cite{Nowruzi}}
\label{fig:hh_stack}
\vspace{-0.2in}
\end{figure*}

The CNN-based approach for homography estimation as
presented in \cite{Nowruzi} attempts to improve on the
accuracy achievable by the previous approach by using
multiple Siamese CNNs stacked end-to-end. Much like the
previous CNN based method, this method also takes for input
the two image patches and outputs the estimated corner
perturbation values $\textbf{H}_{4pt}$, which can be mapped
to a homography matrix $\textbf{H}_{mat}$.

This CNN architecture also uses eight convolutional layers
followed by two fully connected layers (Fig.
\ref{fig:hh_model}). However, the first four convolutional
layers are two parallel branches with shared weights. The
first branch takes the original patch as input, and the
second branch takes the warped patch, as opposed to stacking
the patches and inputting into the same layer. The branches
merge after the fourth layer, concatenating along the
feature dimension. Each layer is followed by ReLu activation
and batch normalization. Max Pooling is applied after the
second, fifth, and seventh convolutional layers. Dropout of
factor 0.5 is applied prior to the first fully connected
layer.

This architecture is repeated through separate modules
stacked end-to-end as shown in Fig. \ref{fig:hh_stack}. The
original patch and the corresponding warped patch are fed
into the first module, which estimates
$\textbf{H}_{4pt}$. That estimated homography is subtracted
from the target to create a new target for the next module.

\begin{equation}\label{Hboost}
\textbf{H}_{i}^{target}  = 
\textbf{H}_{i-1}^{est} -
\textbf{H}_{i}^{target}
\end{equation}

The new target is used to create a new warped patch. The
input data to the following module is the original patch and
new warped patch, with the new target as ground
truth. ($\textbf{I}$ represents the input image patch.)

\begin{equation}\label{Iboost}
(\textbf{I}_{i}^{original},\textbf{I}_{i}^{warped}) =
(\textbf{I}_{i-1}^{original},\textbf{I}_{i-1}^{original} * \textbf{H}_{i}^{target})
\end{equation}

This process is repeated iteratively, with a ``boosting''
like effect that drives the error down. In other words, the
magnitude of the error residuals shrinks between each
iteration. By training each module specific to those error
residual ranges, the overall accuracy is increased.

\section{Dataset Used and the Metrics for the Comparative Evaluation}
\label{sec:datasets_used}

Our study is based on the Open Images 2019 Test Set
(approximately 100k images) \cite{OpenImages}
\cite{OpenImages2}. This is the first published use of CNNs
for homography estimation on the Open Images
Dataset. Additionally, we use the COCO 2017 Unlabeled
Dataset (approximately 123k images)
\cite{Lin2014MicrosoftCC} for CNN training and validation.
By using a different dataset for training/validation and for
testing, we ensured that the CNNs generalized
appropriately.

The primary metric for testing accuracy was Average Corner
Error. This is the Euclidean distance between corner
locations after applying the target and output
homographies. The value was averaged over the four corners.

\begin{equation}\label{eq:ACE}
\begin{split}
    ACE = \frac{1}{4} \sum_{i=1}^4 || \textbf{H}_{mat}^{target} \textbf{x}_i - \textbf{H}_{mat}^{output} \textbf{x}_i || \\
\end{split}
\end{equation}
Here, $\textbf{x}_i$ refers to one of the four corners of
the original patch, $\textbf{H}_{mat}^{target}$ is the
ground truth homography in the original matrix
parameterization, and $\textbf{H}_{mat}^{output}$ is the
estimated homography from the technique being evaluated.

We present median ACE as opposed to mean ACE due to the
unbounded error at the positive extreme. The ACEs of the
highest error greatly skew the mean. This skewing effect
heavily favors CNNs and fails to illustrate which method is
most accurate for the majority of the data.

Previous publications on homography estimation with deep
learning did not report on how the error is distributed over
the dataset for each method
\cite{DBLP:journals/corr/DeToneMR16}
\cite{Nowruzi}. Therefore, in this report we also include a
novel metric named Outlier Ratio (OR). This is the ratio of
extremely high outlier ACE values compared to the rest of
the dataset. For the sake of consistency, \textit{we define
  a value to be an outlier if it exceeds an ACE of 50
  pixels}, or is undefined because the feature-matching
method failed to produce enough correspondences for a
solution. We choose 50 since it is such extreme error that
if a homography maps points that are greater than 50 pixels
away from ground truth in an image patch of size $128 \times
128$, that estimated homography has little practical
use. This OR metric can be heuristically thought of as the
rate at which a certain method will produce a practically
unusable homography in given conditions.

\section{Setting Up the Experiments for Comparative Evaluation}
\label{sec:comparative_eval}

We implemented and analyzed the performance of the two deep
learning solutions under conditions including those of high
variance.  And we have also compared the performance
obtained with the deep-learning based methods with the
baseline method that derive a homography using the interest
points generated by using the SIFT, SURF, and ORB operators.

We execute the homography estimation tests for all methods
on the Open Images 2019 Test Set \cite{OpenImages}
\cite{OpenImages2}. Using this dataset, we conduct four
experiments by simulating different conditions of variance,
including the ideal (i.e. unaltered input images), noise,
illumination shifts, and random occlusions.

In addition to the performance evaluations mentioned above,
we chose the DH method for evaluating the CNN-based
homography estimation when the CNN is trained on noisy data
itself. The DH method was chosen for such experiments
because the architecture is the same as that of the widely
used VGG network, which makes it more likely that the trends
found with DH-based experiments are more likely to be
experienced with other common CNNs. On the other hand, the
HH network uses a Siamese architecture in a sequentially
modular layout, which makes HH harder and longer to train
and makes it more challenging to select the best
hyperparameters for convergence.

We also trained DH on color images (which required making
slight modifications to the network).

All training and validation for CNNs was conducted on the
COCO Unlabeled 2017 Dataset \cite{Lin2014MicrosoftCC}.

As mentioned earlier, the performance was measured with the
Average Corner Error (ACE), as well as the Outlier Ratio
(OR). Finally, we report the results in tables comparing ACE
and OR, as well as figures illustrating the distribution of
sorted ACE values for each method per experiment.

\subsection{Implementation}

As mentioned in the Introduction, in addition to comparing
the two deep-learning method for homography estimation, we
also compare them with the traditional baseline methods that
estimate homographies using the interest points returned by
the SIFT, SURF, and ORB operators.  For these operators, we
used OpenCV 3's \cite{opencv_library} default
implementation.  After determining correspondences, we solve
the homography and refine it with RANSAC.

We implemented and trained the neural networks using the
PyTorch library \cite{pytorch}. 
Labels were created ``on the
fly'' per the method developed in \cite{DBLP:journals/corr/DeToneMR16}
during training and validation. We used the Adam optimizer
\cite{Kingma2014AdamAM} for both networks, training for
approximately 30 epochs at a learning rate of 0.005. (One
epoch here is defined as a single pass over the entire
training dataset.) The learning rate was halved every five
epochs. PyTorch's \verb'MSELoss' function was used as the
loss function.

\subsection{Evaluation Experiments}

We conducted four experiments over the entire dataset. Each
experiment simulates a different form of variance
experienced in natural conditions. The four experiments are \textbf{Ideal Conditions} (where the dataset is unaltered), \textbf{Gaussian Noise}, \textbf{Illumination Shift}, and \textbf{Occlusion}.
For the latter three experiments, the process is repeated
three times with different variance values to
better illustrate the sensitivity of each method as the
variance increases. Table \ref{tab:expriment_breakdown}
breaks down each experiment with the variance values.

To simulate noise, we added normal Gaussian noise scaled by
a factor $\eta$ (such as 0.4) relative to pixel range, and
clipped the results to normalized pixel range.

\noindent
\begin{equation}\label{eq:noise}
\begin{split}
    X_{noisy} = min( max( X + \eta \mathcal{N}(0, 1), -1) , 1)
\end{split}
\end{equation}

To simulate illumination, we multiplied the pixel values in
the warped patch by a factor $\lambda$ (such as 1.4) and
clip the results to normalized pixel range.

\noindent
\begin{equation}\label{eq:illum}
\begin{split}
    X_{illum} = min( max( \lambda X, -1 ) , 1 )
\end{split}
\end{equation}

To simulate occlusions, We replaced an $n \times n$ box of
random location with a random color in the warped image. The
size of $n \times n$ is determined by a factor $\alpha$
(such as 0.6) relative to the total patch space.

\begin{table}
    \centering
    \begin{center}
    \begin{tabular}{|l||c|c|c|}
        \hline
        Experiment & \multicolumn{3}{|c|}{Magnitude of Variance} \\ \hline \hline
        Ideal Conditions & \multicolumn{3}{|c|}{\textit{No Variance}}\\ \hline
        Gaussian Noise & $\eta = 0.1$ & $\eta = 0.3$ & $\eta = 0.5$ \\ \hline
        Illumination Shift & $\lambda = 1.2$ & $\lambda = 1.4$ & $\lambda = 1.6$ \\ \hline
        Occlusion & $\alpha = 0.2$ & $\alpha = 0.4$ & $\alpha = 0.6$ \\
        \hline
    \end{tabular}
    \end{center}
    \caption[Experiment Summary]{Experiments for
      investigating the performance of CNN-based homography
      estimation under four types of variance.  In each
      experiment (except for Ideal), three different magnitudes of variance are
      used to better illustrate the sensitivity of each
      method.}
    \label{tab:expriment_breakdown}
\vspace{-0.2in}
\end{table}

After comparing CNNs to the baseline feature matching
methods, we analyzed the affect of training on noisy data
will have on performance. We used the trained Deep Image
Homography Net, train it for an additional 30 epochs in
$\eta = 0.1$ noisy data, and then measure performance. We
then train it for another 30 epochs on $\eta = 0.3$, tested
again, and then repeat with $\eta = 0.5$.

Finally, to measure the effect of using color images instead
of grayscale images for input, we modified the Deep Image
Homography Net to accept six channels for input instead of
two. We then train the network from scratch on color images,
and measure performance against the grayscale counterpart.

\section{Results of Comparative Evaluation}
\label{sec:results}

We illustrate our results with tables depicting the median
ACE and OR values for each method under a specific
variance. We also present figures that compare the sorted
ACE values as line charts for each method in each
experiment. This best illustrates the distribution of error
over the dataset, and therefore conveys a qualitative sense
of consistency. The lines depict sorted ACE values, from
least to greatest, for each input patch pair. For brevity,
DH refers to the CNN based method described in Section
\ref{sec:dhcnn} and HH refers to the CNN based method
described in Section \ref{sec:hhcnn}.  In the tables below,
the best ACE is in bold. The methods above the dotted line
are CNNs, and the methods below are feature matching.

\subsection{Ideal Conditions}

Under ideal conditions, meaning that when the dataset is
unaltered, as shown in Table \ref{tab:unaltered}, SIFT gives
superior performance for 90\% of the dataset. It has
sub-pixel ACE for approximately half the dataset. HH is the
most consistent, remaining below an ACE of 50 for over 99\%
of the dataset.

\begin{table}[!htb]
\vspace{-0.1in}
\begin{center}
\begin{tabular}{|l|c|c|}
\hline
Method & Median ACE & OR\\
\hline
DH & 3.97 & 0 \\
HH & 2.25 & 0.01 \\ \hdashline
SIFT & \textbf{0.89} & 0.08 \\
SURF & 3.15 & 0.09 \\
ORB & 7.82 & 0.12 \\
\hline
\end{tabular}
\centering
\caption[Median ACE and OR in Ideal Conditions]{Median ACE and OR in Ideal 
  Conditions. SIFT has the
  lowest median ACE, followed by HH. 
  ORB was the
  worst-performing method, with a high median ACE and OR.}
\label{tab:unaltered}
\vspace{-0.25in}
\end{center}
\end{table}

\begin{figure}[!htb]
\vspace{-0.1in}
\begin{center}
    \includegraphics[width=0.8\linewidth]{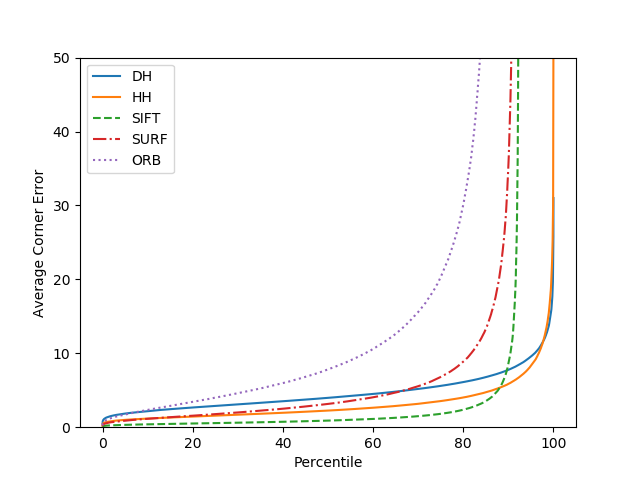}

   \caption[Sorted ACE for Ideal Conditions.]{Sorted ACE for Ideal Conditions. SIFT gives superior performance for 90\% of the data. However, in the final 8\% of the dataset, SIFT has incredibly high error.}
\label{fig:unaltered}
\vspace{-0.3in}
\end{center}
\end{figure}

Fig. \ref{fig:unaltered} shows the plots of the ACE values
achieved with all five approaches to homography estimation,
the three traditional methods based SIFT, SURF, and ORB, and
the two CNN based methods. As is evident from the plot, the
feature-matching approach based on SIFT produces the most
accurate homographies for 90\% of the dataset.  However,
when it fails for the rest of the dataset, it fails
miserably.  On the other hand, the CNN based solution have
slightly worse accuracies, but produce usable results for a
larger fraction of the dataset.

\subsection{Noise}

As shown in Table \ref{tab:noise}, the traditional
feature-matching based methods are much more sensitive to
noise than the CNN based methods.  The SIFT and the SURF
based homography estimators maintain a lower median ACE
value than the CNN-based methods at $\eta = 0.1$. However,
when $\eta = 0.3$, both the SIFT and the SURF based methods
fail to find a sufficient number of feature correspondences
for reliable homography estimation for over half the
dataset.  The two CNN method, though obviously affected by
noise, remain much more robust over a significant fraction
of the dataset.

\begin{table}[!htb]
\begin{center}
\begin{tabular}{|l|c|c|c|}
\hline
Noise $\eta $& 0.1 & 0.3 & 0.5 \\
\hline\hline
Method & \multicolumn{3}{|c|}{Median ACE / OR} \\ \hline
DH & 13.60 / 0.01 & 21.73 / 0.01 &  28.49 / 0.03\\
HH & 6.79 / 0.01 & \textbf{13.38} / \textbf{0.01} & \textbf{17.30} / \textbf{0.01} \\ \hdashline
SIFT & \textbf{2.33} / 0.18 & NAN / 0.66 & NAN / 0.94\\
SURF & 5.42 / 0.17 & NAN / 0.62 & NAN / 0.93 \\
ORB & 12.20 / 0.16 & 26.50 / 0.34 & 53.12 / 0.52 \\
\hline
\end{tabular}
\end{center}
\caption[Median ACE and OR in Gaussian Noise]{Median ACE and
  OR value in the presence of additive noise.
  HH has the best median ACE for
  ``moderate'' and ``high'' noise levels. ``NAN'' 
  stands for ``Not a Number''.}
\label{tab:noise}
\vspace{-0.1in}
\end{table}

\begin{figure}[!htb]
\vspace{-0.1in}
\begin{center}
    \includegraphics[width=0.8\linewidth]{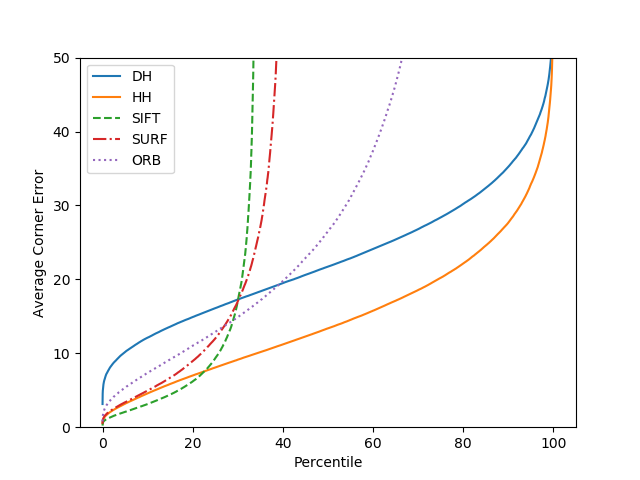}
\end{center}
   \caption[Sorted ACE for Noise $\eta = 0.3$]{Sorted ACE
     for the noise level $\eta = 0.3$. At a 
     ``moderate'' level of noise, SIFT and SURF fail to give viable results for
     over 60\% of the dataset. ORB is
     more robust, but not as robust as the CNNs. }
\label{fig:noise_30}
\vspace{-0.2in}
\end{figure}

Fig. \ref{fig:noise_30} shows the plots for the case of
additive noise at the additive level of $\eta = 0.3$.  It is
interesting to see while the traditional approach to
homography estimation based on SIFT gives the best results
on a small fraction of the dataset, it essentially fails to
produce any usable results at all for a large part of the
dataset.  The other two traditional approaches do worse than
the one based on SIFT.  On the other hand, the two CNN based
approaches work better for a much larger fraction of the
overall dataset, although not matching the best (in terms of
accuracy) that one can get with a SIFT based method.

\subsection{Illumination}

As shown in Table \ref{tab:illum}, the traditional
feature-matching based methods have a better median ACE at
``low'' and ``moderate'' levels of shifts in illumination,
but the HH based CNN homography estimator shows better
Between the two CNNs,
DH is impacted much more significantly by illumination
shift.  The discrepancy between the two CNNs could be
explained by the fact that HH is based on a Siamese network,
where two input patches are transformed by four
convolutional layers before the features are combined.  By
contrast, DH stacks the input patches at the very beginning.
The discrepancy in input patches, therefore, has a larger
impact, as there is no opportunity for the network to
isolate invariant features within each patch, independent of
the other.

\begin{table}[!htb]
\begin{center}
\begin{tabular}{|l|c|c|c|}
\hline
Illum $\lambda$ & 1.2 & 1.4 & 1.6 \\
\hline\hline
Method & \multicolumn{3}{|c|}{Median ACE / OR} \\ \hline
DH & 23.08 / 0.01 & 26.45 / 0.01 & 28.38 / 0.01 \\
HH & 2.35 / 0.01 & 2.86 / 0.01 & \textbf{5.50} / 0.01 \\ \hdashline
SIFT & \textbf{0.95} / 0.10 & \textbf{1.35} / 0.17 & 10.70 / 0.47 \\
SURF & 3.41 / 0.12 & 5.65 / 0.26 & NAN / 0.67 \\
ORB & 8.48 / 0.18 & 11.92 / 0.26 & 48.26 / 0.50 \\
\hline
\end{tabular}
\end{center}
\caption[Median ACE and OR in Illumination Shift]{Depicted
  are the ACE and OR for each method for the caes of
  illumination shifts. SIFT has a lower median ACE at
  ``low'' and ``moderate'' levels.
  At ``high'', HH has a lower median ACE.}
\label{tab:illum}
\vspace{-0.1in}
\end{table}

\begin{figure}[!htb]
\vspace{-0.15in}
\begin{center}
    \includegraphics[width=0.8\linewidth]{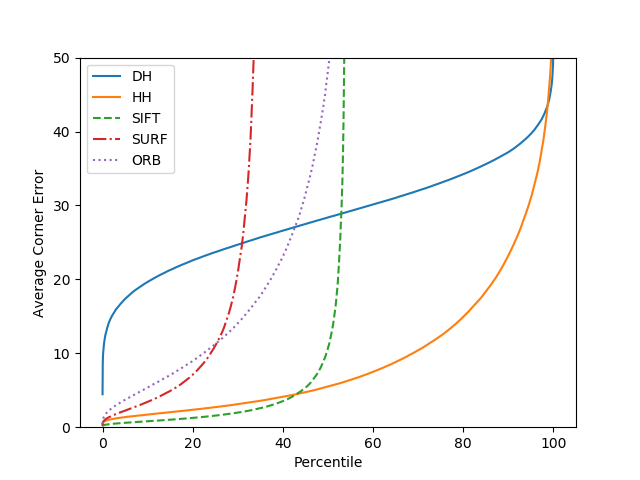}
\end{center}
  \caption[Sorted ACE for Illumination $\lambda = 1.6$]{Sorted ACE for Illumination $\lambda = 1.6$.  HH has the lowest ACE for 56\% of the dataset.}
  \label{fig:illum_60}
\vspace{-0.2in}
\end{figure}

Fig. \ref{fig:illum_60} shows how the performance varies
across the dataset for illumination
variance at $\lambda = 1.6$. 
Under conditions of illumination variance, the traditional
approaches to homography estimation work for only a small
fraction of the dataset. 
Of the two CNN based methods, the DH solution does
miserably across the board.  While DH does not pass muster
for this difficult case, what is achieved by the HH solution
is indeed impressive.

\subsection{Occlusion}

In comparison with additive noise and illumination variations, it is
interesting to see that the traditional feature-matching
based methods for homography estimation are not as sensitive
to occlusions. As shown in Table \ref{tab:occluded}, the
SIFT based homography estimator retains the lowest median
ACE at all three levels of occlusion while the two CNN based
methods remain more consistent. This suggests that, as long
as there is enough opportunity for distinct key points to be
identified and matched, feature matching will experience
little degradation.

\begin{table}[!htb]
\begin{center}
\begin{tabular}{|l|c|c|c|}
\hline
Occlude $\alpha$ & 0.2 & 0.4 & 0.6 \\
\hline\hline
Method & \multicolumn{3}{|c|}{Median ACE / OR} \\ \hline
DH & 4.74 / 0.01 & 6.99 / 0.01 &  11.23 / 0.01 \\
HH & 2.27 / 0.01 & 2.39 / 0.01 & 3.01 / 0.01 \\ \hdashline
SIFT & \textbf{0.91} / 0.10 & \textbf{0.99} / 0.13 & \textbf{1.29} / 0.21 \\
SURF & 3.52 / 0.12 & 4.85 / 0.20 & 14.94 / 0.41 \\
ORB & 9.75 / 0.21 & 22.04 / 0.39 & 96.03 / 0.74 \\
\hline
\end{tabular}
\end{center}
\caption[Median ACE and OR in Occluded Conditions]{Values
  for the Median ACE and OR under occlusion.  SIFT has the
  best median ACE for the different degrees of occlusion and
  experiences only a minor degradation of OR at $\alpha =
  0.6$.}
\label{tab:occluded}
\vspace{-0.15in}
\end{table}

\begin{figure}[!htb]
\vspace{-0.15in}
\begin{center}
    \includegraphics[width=0.8\linewidth]{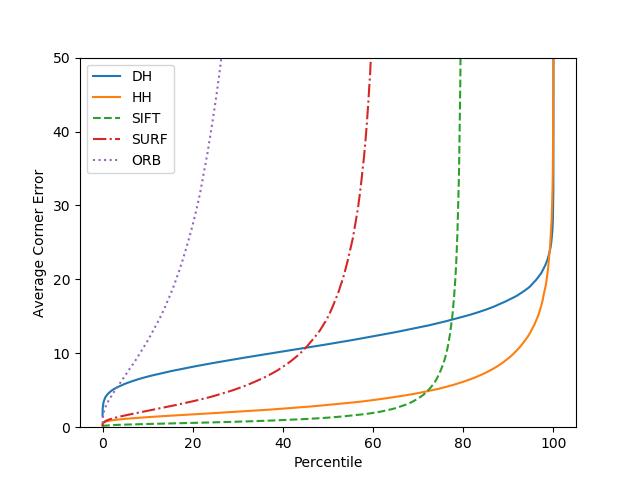}
\end{center}
  \caption[Sorted ACE for Occlusion $\alpha = 0.6$]{Sorted
    ACE values for the occlusion level of $\alpha = 0.6$.}
  \label{fig:occlude_60}
\vspace{-0.2in}
\end{figure}

Fig. \ref{fig:occlude_60} shows the performance plots for
all five cases.  SIFT does the best compared to the other two traditional methods.  
And, what is even more interesting, for a significant fraction of
the images, SIFT even beats out the two CNN-based methods.
But, as can be inferred from the rapid rise of the SIFT
plot, it fails miserably at 79\%.  On the other
hand, the two CNN based methods fail completely
at a much smaller rate than SIFT.

\subsection{Color vs. Grayscale}

As shown in Table \ref{tab:dh_color}, training the DH CNN
with color inputs instead of grayscale inputs resulted in a
slightly larger median error for the ACE metric.

\begin{table}[!htb]
\vspace{-0.1in}
\begin{center}
\begin{tabular}{|l|c|c|}
\hline
Method & Median ACE & OR\\
\hline
DH Grayscale & 3.97 & 0 \\
DH Color & 4.74 & 0 \\
\hline
\end{tabular}
\centering
\caption[Comparison of DH Trained in Grayscale to DH Trained
  in Color]{Comparison of the DH network trained with
  grayscale images.}
\label{tab:dh_color}
\vspace{-0.3in}
\end{center}
\end{table}

\begin{figure}[ht]
\vspace{-0.15in}
\begin{center}
    \includegraphics[width=0.8\linewidth]{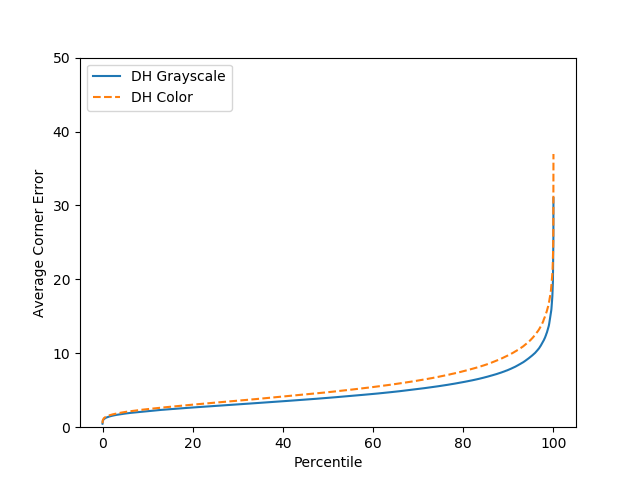}
\end{center}
  \caption[DH Trained In Grayscale vs DH Trained In
    Color.]{Performance plots for the color vs. grayscale
    comparison. Contrary to intuition, the results with
    color are slightly worse.}
\label{fig:dh_color}
\vspace{-0.2in}
\end{figure}

The performance plots are shown in Fig. \ref{fig:dh_color}.
The graph for the grayscale
case is consistently below the graph for color, implying
that, with the same CNN architecture, the
grayscale images result in slightly more accurate
homographies than the color images.

\subsection{DH Trained with Noise-Corrupted Inputs}

Table \ref{tab:dh_noise} presents the results obtained when
the DH is trained with different amounts of noise added to
the training dataset.  
The CNN based homography estimator achieved significantly
superior performance when trained to the specific level of noise.  
However, this performance enhancement came with a small degradation in the performance for the noiseless
case (Ideal).  Additionally, training at a lower level of
noise only slightly improved the performance at higher
levels of noise.

\begin{table}[!htb]
\vspace{-0.1in}
\begin{center}
\begin{tabular}{|l|c|c|c|c|}
\hline
Noise $\eta$ & 0 & 0.1 & 0.3 & 0.5 \\
\hline\hline
Method & \multicolumn{4}{|c|}{Median ACE} \\ \hline
DH & \textbf{3.97} & 13.60 &  21.73 & 28.49 \\
DH Noise 10 & 10.08 & \textbf{4.24} &  20.93 & 26.40 \\
DH Noise 30 & 6.25 & 5.99 & \textbf{4.99} & 20.17 \\
DH Noise 50 & 5.66 & 5.54 & 5.62 & \textbf{5.70}\\
\hline
\end{tabular}
\end{center}
\caption[Deep Image Homography Trained With Noise.]{Deep Image Homography Trained With Noise. The
  table shows the median ACE values for the case when the DH
  network is trained with different degrees of noise.}
\label{tab:dh_noise}
\vspace{-0.1in}
\end{table}

\begin{figure}[ht]
\vspace{-0.15in}
\begin{center}
    \includegraphics[width=0.8\linewidth]{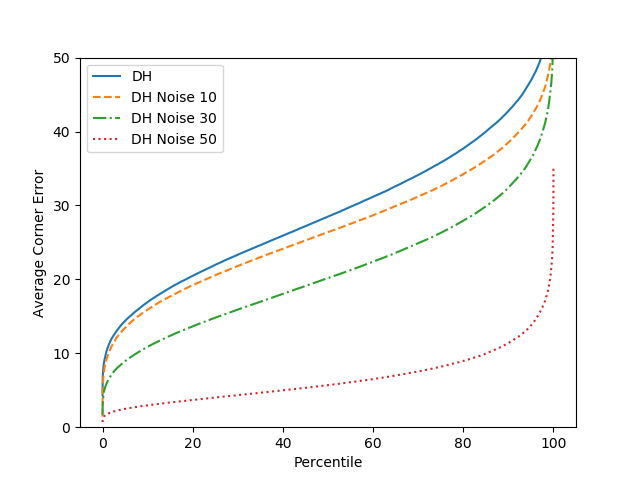}
\end{center}
  \caption[Sorted ACE values for DH Trained to Noise in
    Noise $\eta = 0.5$ Dataset.]{Sorted ACE values for the
    DH network that is trained to noise with the $\eta =
    0.5$ dataset. The CNN trained with the highest level of
    noise also performs the best under high noise
    conditions.}
\label{fig:dh_noise_50}
\vspace{-0.25in}
\end{figure}

Shown in Fig. \ref{fig:dh_noise_50} are the percentile plots
for the case when the three different DH networks, each trained with one
of the three noisy input datasets as listed in Table
\ref{tab:dh_noise}), are tested on the same input-plus-noise
data set corresponding to $\eta = 0.5$.  Since the
red-dotted plot for the case ``DH Noise 50'' is the slowest
rising graph in the figure, this again shows that one gets
the best results when the level of noise at which the DH is
trained matches the level of noise in the input images.

\section{Conclusion}
\label{sec:conclusion}

Under what we have referred to as the Ideal conditions and
under conditions of low variance, feature matching based
homography estimation (particularly SIFT) is still the most
accurate method (per median ACE).

However, in conditions of moderate to heavy variance,
especially with Gaussian noise, CNN-based homography
estimation achieves a lower median ACE. 
Under these
conditions, the feature matching based
methods consistently suffered sensitivity to variance.
The CNN-based homography estimators learned to be more
robust even when the training was confined to ideal
conditions.

Our experiments have also demonstrated that the robustness
to noise for CNN-based homography estimators can be
further improved by training with noisy data.  Our
experiments showed that the performance improved
significantly at the specific level of noise with which the
CNN trained. However, that improved performance 
comes at a slight degradation to accuracy in ideal
conditions. There was only little improvement to performance in higher magnitudes of noise. 
Essentially, by training
to a specific magnitude of noise, a ``Goldilocks Zone'' is
created where that CNN performs best. More experiments need
to be conducted to determine the ideal process for training
in noise that optimizes performance across all potential
magnitudes of noise.

CNN-based homography
estimators produced more consistent results than the
feature-matching based estimators in all conditions. 
While SIFT produced outputs with unacceptable errors on 8\% of the
Ideal conditions, the CNN based methods produced
such outputs for less than one percent in any conditions. 
This is likely because
CNNs can be trained to a specific range of homography
values, unlike feature-matching. Many applications, such as real-time tracking for AR,
might benefit in consistency over absolute error.

Explicitly including color channels at the input for
training CNN based homography estimators did not produce
significantly better results for CNN based methods. Despite
the additional information provided by the different color
channels, the CNN-based homography estimation actually
performed slightly worse. 
However, it is possible that a more
elaborate architecture
would yield higher accuracies for the homographies estimated
from color images.  Additionally, given the redundancy
between the RGB channels of a color image, such inputs might
increase robustness
against noise. More research is necessary to find the
optimal architecture and training methods that best
leverages the additional information available in color images.

The overall lesson learned from our research is that no one
method for homography estimation can be treated as
``universally superior'' to all others.  The environmental
conditions and engineering constraints need to be
deliberately considered when choosing the most appropriate
technique in any application. While CNN-based homography
estimators may not be as accurate as the best
``hand-crafted'' feature-based algorithms, they are more
consistent and perform better in conditions of variance
(noise, illumination shifts, and occlusions). 


{\small
\bibliographystyle{ieee}
\bibliography{egbib}
}

\end{document}